\documentclass{ecai} 

\usepackage{latexsym}
\usepackage{amssymb}
\usepackage{amsmath}
\usepackage{amsthm}
\usepackage{booktabs}
\usepackage{enumitem}
\usepackage{graphicx}
\usepackage{color}
\usepackage[table,xcdraw]{xcolor}

\newcommand{\BibTeX}{B\kern-.05em{\sc i\kern-.025em b}\kern-.08em\TeX}
\newcommand{\net}{StyleMamba }

\begin{document}


\begin{frontmatter}


\paperid{2447} 


\title{\net: State Space Model for Efficient Text-driven Image Style Transfer}


\author[A,C]{\fnms{Zijia}~\snm{Wang} \thanks{Email: zijia.wang18@imperial.ac.uk}}
\author[B]{\fnms{Zhi-Song}~\snm{Liu}}

\address[A]{Imperial College London, the United Kingdom}
\address[B]{LUT School of Engineering Sciences, Finland}
\address[C]{Dell Research UK, the United Kingdom}

\begin{abstract}
We present \net, an efficient image style transfer framework that translates text prompts into corresponding visual styles while preserving the content integrity of the original images. Existing text-guided stylization requires hundreds of training iterations and takes a lot of computing resources. To speed up the process, we propose a conditional State Space Model for Efficient Text-driven Image Style Transfer, dubbed \net, that sequentially aligns the image features to the target text prompts. To enhance the local and global style consistency between text and image, we propose masked and second-order directional losses to optimize the stylization direction to significantly reduce the training iterations by 5$\times$ and the inference time by 3$\times$. Extensive experiments and qualitative evaluation confirm the robust and superior stylization performance of our methods compared to the existing baselines.
\end{abstract}

\end{frontmatter}


\section{Introduction}

In recent years, the intersection of computer vision and natural language processing has led to significant advancements in multimodal perception and understanding. One particularly intriguing area of research within this domain is text-driven or text-guided image style transfer. This emerging field explores the synthesis of images guided by textual descriptions, enabling the transformation of visual content to match desired artistic styles, scenes, or aesthetics as described in natural language. This fusion of vision and language offers promising avenues for creative expression and content generation.
\begin{figure*}[t]
    \centering
    \includegraphics[width=\linewidth]{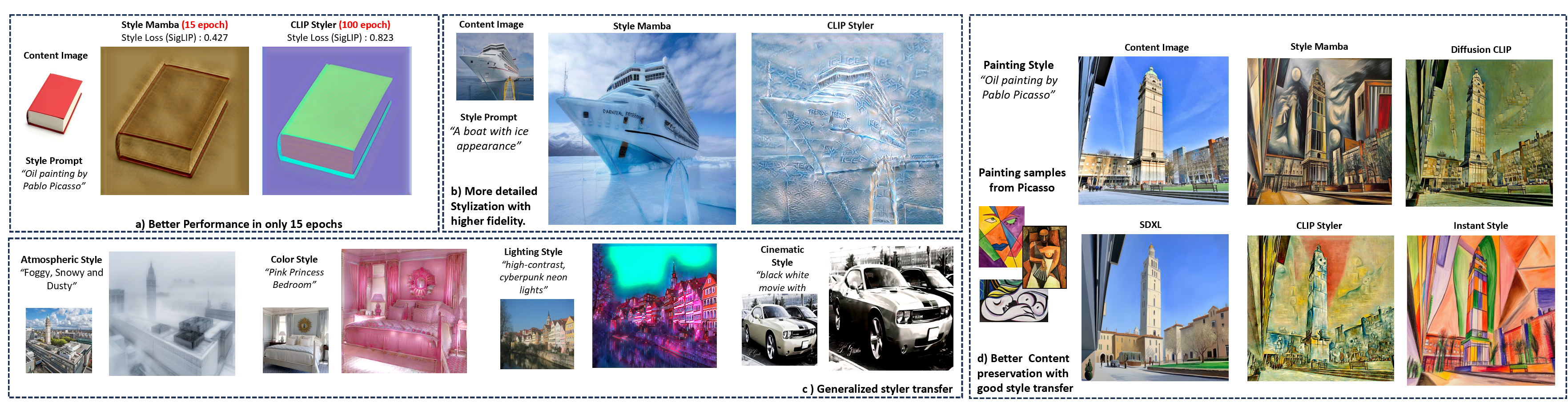}
    \caption{\textbf{Comparative results showcasing the efficacy of the \net framework.} a) highlights the rapid convergence and stylization capabilities with fewer of epochs. b) and c) demonstrate the detailed stylization fidelity and the transferability of various styles, including painting and lighting styles, as well as some complicated styles like environmental style. Finally, d) shows the great content preservation ability of \net with competitive style transfer performance compared with other style transfer models. All results reflect the superiority of \net.}
    \label{fig:teaser}
    \vspace{10pt} 
\end{figure*}

Unlike image-driven style transfer, one notable advantage of text-driven style transfer is its enhanced flexibility and interpretability. Texts provide a more abstract and semantically rich representation of desired styles or attributes than explicit reference images. This abstraction enables users to express complex artistic concepts and emotions that are challenging to convey through images alone. Existing text-driven approaches, e.g., Clipstyler~\cite{clipstyler}, DiffusionCLIP~\cite{diffusionclip} and TxST~\cite{txst}, show tremendous promising results in visual diversity. The key idea is to align the text and image embeddings~\cite{clip,siglip}, such that the stylized image can be faithful to the direction of the text. However, they need to consider the nuances of style-specific text description and visual style consistency. More importantly, they require hundreds of training iterations and considerable GPU resources, which is inefficient for real applications. 

Thus, we introduce \net, a framework to incorporate a conditional State Space Model~\cite{mamba} into the AutoEncoder systems which can use text prompts to supervise the style fusion process. It can significantly improve the training converge with $10\sim 20\times$ speedup. Furthermore, we explore the Masked and Second-order directional loss to better align text and style. Figure~\ref{fig:teaser} showcases the superiority of \net in performing style transfer, illustrating its capability to rapidly assimilate and apply complex styles to various content images with remarkable fidelity. Through a series of comparisons, \net demonstrates enhanced performance in not only fewer epochs  (Figure ~\ref{fig:teaser} a) but also finer detail preservation  (Figure ~\ref{fig:teaser} b) and adherence to diverse artistic prompts(Figure ~\ref{fig:teaser} c), ranging from mimicking the thick textures of Picasso’s oil paintings to capturing the essence of atmospheric and lighting conditions. In Figure ~\ref{fig:teaser} d, these examples highlight \net’s better balance between content preservation and style transfer ability, enabling more generalized applications. Our contributions can be summarized below:

\begin{itemize}
	\item We propose a simple framework, \net, incorporating the conditional Mamba into AutoEncoder to achieve fast text-driven style transfer (Section~\ref{sec:framework}).
	\item Our proposed framework achieves better stylization via novel Masked directional loss and Second-order relational loss (Section~\ref{sec:direction_loss}), which can speed up overall stylization and better grasp global and local style consistency without compromising the contents. 
	\item We empirically show the efficiency of the proposed \net over the existing state-of-the-art techniques in both quantitative and qualitative measurements. Owing to the simplistic nature of \net, we further experiment with video style transfer and multiple style transfer to show the versatility of our framework.
\end{itemize}

\section{Related Work}
\label{sec:related_work}


\noindent \textbf{Image-driven style transfer.} Style transfer aims to fuse desirable style to the content images such that the resultant image can preserve the contents while showing the desired style, i.e., color and textures. Early approaches such as Gatys et al.'s work~\cite{Gatys} on neural style transfer laid the groundwork by formulating style transfer as an optimization problem, leveraging deep convolutional neural networks to separate content and style representations within images. Building upon this foundation, subsequent research~\cite{adaattn,adain,mgad,sanet,vaest,cast} has explored various techniques to improve the efficiency, stability, and fidelity of style transfer algorithms. For instance, Li et al.~\cite{adain} introduced adaptive instance normalization (AdaIN), which dynamically adjusts the statistics of intermediate feature maps to better match the style of a reference image. Other approaches have incorporated perceptual loss functions inspired by human visual perception to preserve semantic content during style transfer better.  Additionally, recent works~\cite{artflow,dstn,quantart,wise} have explored conditional and controllable style transfer, allowing users to specify desired style attributes or manipulate style transfer outcomes with good visual quality. 

\noindent \textbf{Text-image multimodality.} Benefiting from the study of LLMs and VLMs, it becomes popular to use contrastive learning~\cite{simclr,simclr2} to align general texts and images. CLIP~\cite{clip} is one of the pioneering works that can use natural language supervision for image representation. The key idea is to utilize pre-trained language and image encoders to project texts and images onto the compressed domain and maximize cosine similarity between paired text-image data. To further improve the generalization and text-image alignment, SigLIP~\cite{siglip} proposes to replace conservative learning with pairwise sigmoid loss. It can scale up the batch size and end up with memory-efficient and superior text-image alignment. Both CLIP and SigLIP offer zero-shot capacities that can be used for many text-image multimodal tasks. For example, \cite{unclip} proposes a text-conditional image generation that can convert texts to desirable images. The Stable Diffusion Model~\cite{ldm} trains a CLIP-based autoencoder that can take text embeddings as conditions to further improve image quality. A similar concept can also be applied to video generation~\cite{clip4clip,language_6}, video retrieval~\cite{clipvit}, 3D generation~\cite{dreamgaussian,magic3d}, 3D editing~\cite{3dedit} and image captioning~\cite{expressive}.

\noindent \textbf{Text-driven style transfer.} Different image-driven style transfer, text-driven style transfer uses text prompts to guide the stylization process. Clipstyler~\cite{clipstyler} is the first work to utilize CLIP to project both stylized images and texts to the latent space. It can use pairwise cosine similarity to maximize their distance. It further inspires several works on improving quality or speeding up the training iterations. For example, DiffusionCLIP~\cite{diffusionclip} combines the state-of-the-art pre-trained diffusion model~\cite{ldm} and CLIP to learn stylization in the latent space. Gatha~\cite{gatha} proposes to modify the global directional loss between texts and images as a relational loss, which is calculated based on a style tensor that can question whether the generated image is truly aligned with the target texts. One of the key disadvantages of the aforementioned methods is the online training process. It usually takes up to 150 iterations (approximately 10 minutes on one standard GPU) to obtain one good stylized image. Recent works either speed up the whole training times~\cite{fastclipstyler} or perform offline contrastive training~\cite{ldast,txst,stylediffusion,zecon}.

\section{Approach}
\label{sec:approach}

\begin{figure*}[t]
    \centering
    \includegraphics[width=\linewidth]{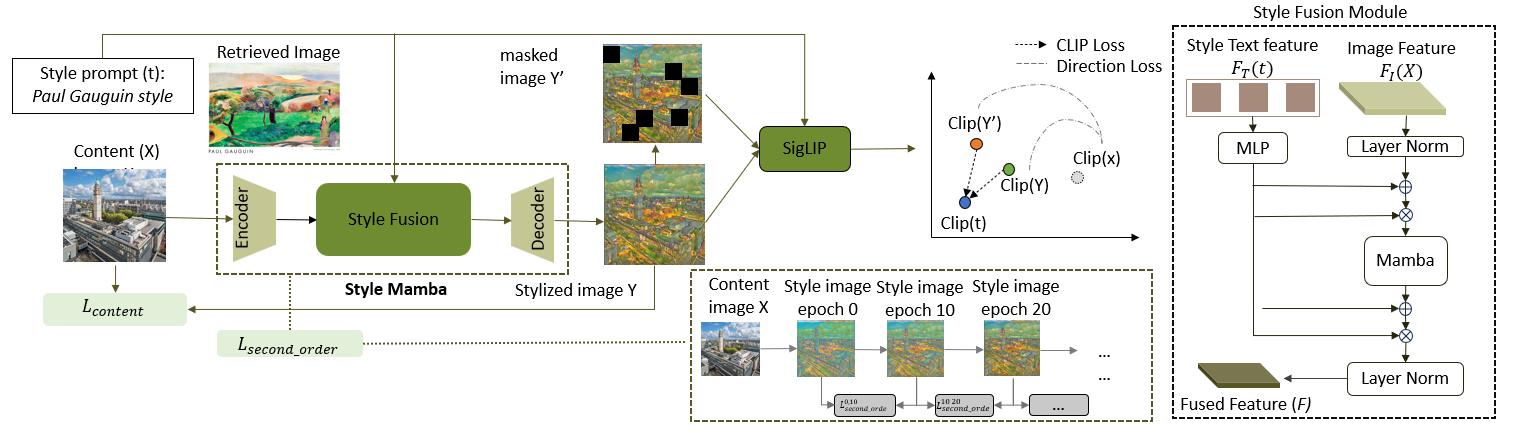}
    \caption{\textbf{Workflow overview of \net framework.} The process begins with a content image and a style prompt (e.g., ``\textit{Paul Gauguin style}''). An encoder converts the content image into a latent representation, which undergoes style fusion with features derived from the style prompt. This fusion is facilitated by the Style Fusion Module, incorporating masked and second-order directional losses to guide the text-to-image stylization. The result is a stylized image $\textbf{Y}$, which closely adheres to the style prompt while preserving content integrity.}
    \vspace{10pt}
    \label{fig:overview}
\end{figure*}

The overall flow of the proposed \net is shown in Figure \ref{fig:overview}. The input content image $\textbf{X}$ is converted to latent representations $F_{e}(\textbf{X})$ using a pretrained Variational AutoEncoder (VAE) $F_{e}(\cdot)$from stable diffusion~\cite{ldm}, while the style text $t$ is converted into embeddings $F_{T}(t)$ using pretrained text encoder $F_{T}(\cdot)$ in SigLIP~\cite{siglip}. They are then fused in the proposed style fusion module to get a new feature map $\textbf{M}$. Guided by the text-to-image style loss, the decoder $F_{d}(\cdot)$ decodes the $\textbf{M}$ into a stylized image $\textbf{Y}$. The training process is guided by the content loss ($\mathcal{L}_{\text{content}}$) based on VGG \cite{VGG} or lpips \cite{LPIPS}, the second order loss ($\mathcal{L}_{\text{so}}$), the masked directional loss ($\mathcal{L}_{\text{md}}$) and global directional loss ($\mathcal{L}_{\text{dir}}$).

\subsection{Overall framework}
\label{sec:framework}

The whole architecture consists of three components: i) \textbf{Auto Encoder}, which is used to encode the content image $\textbf{X}$ and reconstruct the stylized Image $\textbf{Y}$; ii)\textbf{ Style Fusion Module},  which works as a fuser to selectively attend the  $F_{T}(t)$ and  $F_{e}(\textbf{X})$ to derive $\textbf{M}$; iii) \textbf{SigLIP Module}, which is used to encode the style features contained in $F_{T}(t)$ and to act as a 'North Start' to guide the whole process. 

\noindent \textbf{Auto Encoder.}  The proposed masked directional loss, coupled with second-order loss, needs the powerful content-preserving ability of the Auto Encode, while the better reconstruction ability is where the pretrained VAE from the Stable Diffusion Model \cite{ldm} excels, as it is adept at encoding the content image into a rich, compressed latent space while ensuring that the essential features are retained for accurate reconstruction. Moreover, the absence of shortcuts between the encoder and decoder in this architecture means that the stylization process relies entirely on the latent representation, fostering a more profound style integration without residual information from the input image. This approach aligns with the Information Bottleneck principle, striking an optimal balance between data compression and preserving relevant content features for the task. Therefore, the pretrained VAE in Stable Diffusion Model~\cite{ldm} is chosen here to be  $F_{e}(\cdot)$. 

\noindent \textbf{Style Fusion Module.} Inspired by DiT~\cite{dit}, we propose a conditional State Space Model for style fusion. Concretely, the Style Fusion Module in the \net framework effectively combines the textual style features with the content image features by incorporating elements of Adaptive Layer Norm  (AdaLN) and Mamba process \cite{mamba}. Mamba's selective process accelerates the style transfer process by focusing only on relevant features, which speeds up training and inference. AdaLN's ability is to adaptively adjust style characteristics from textual descriptions using learnable shifting and scaling values. The AdaLN approach allows for the adjustment of the normalized content features to resonate with the stylistic elements described in the text, achieving adaptive style manipulation. Specifically, the Style Fusion Module incorporates AdaLN by adjusting feature distributions based on the style features derived from the text, ensuring the content image's integrity while infusing the desired style. The normalization of content feature $F_{e}(\textbf{X})$ is depicted mathematically as:
\begin{equation}
    \textbf{M} = (LN(F_{e}(\textbf{X}) + \alpha_1 \cdot SSM(LN(F_{e}(\textbf{X}))) \cdot \mu_1 + \sigma_1)) + \alpha_2 + \sigma_2
    \label{eq:ssm}
\end{equation}

\noindent In Equation~\ref{eq:ssm}, \( LN \) denotes Layer Normalization, and \( SSM \) represents the State Space Model. In our case, we use Mamba \cite{mamba}, \( \alpha \) and \( \sigma \) are scaling and shifting parameters, respectively, which are learned from the style features. The Mamba, with its selective state space model, allows for content-dependent transformation, selectively attending to relevant information and dynamically adjusting to the content based on the current context. This allows for efficient and adaptive style transfer while maintaining content consistency.

\noindent \textbf{SigLIP Module.} For the SigLIP Module model section, the introduction of the pretrained SigLIP \cite{siglip} is central to enhancing the style fusion process, which is critically dependent on the rich semantic encoding capabilities of the pretrained CLIP models. SigLIP achieves better zero-shot classification accuracy on ImageNet \cite{imagenet_cvpr09} than CLIP \cite{clip}. SigLIP can better generalize to unseen image categories based on natural language descriptions. To demonstrate its superior performance over CLIP, we conduct ablations in the Experiment section.

\subsection{Proposed Text-driven style losses}
\label{sec:direction_loss}

\noindent \textbf{Global directional loss.} In the proposed net framework, the global directional loss $\mathcal{L}_{\text{dir}}$ is the key guidance in aligning the transformation of the image content with the specified text-driven style. Like the direction loss in \cite{clipstyler}, it operates by computing the cosine similarity between the direction vectors of text and image features, ensuring that the stylization process is coherent with the textual description. Formally, this loss is defined as:


\begin{equation}
\begin{matrix}
    \!\begin{aligned}
	& \mathcal{L}_{\text{dir}} = 1 - \frac{\mathbf{T}_{\text{dir}} \cdot \mathbf{I}_{\text{dir}}}{\|\mathbf{T}_{\text{dir}}\| \|\mathbf{I}_{\text{dir}}\|}~\text{, where}\\
	& \mathbf{T}_{\text{dir}} = F_{T}(t) - F_{T}(t_{src}), \quad \mathbf{I}_{\text{dir}} = F_{I}(\textbf{Y}) - F_{I}(\textbf{X})
	\label{eq:Equation2}
	\end{aligned}
\end{matrix} 
\end{equation}

\noindent where $F_{T}$ and $F_{I}$  are the text encoder and image encoder of the SigLIP, respectively, then $t$ and $src$ are style text and source text (here we define it as "a plain photo"), $\mathbf{T}_{\text{dir}}$ and $\mathbf{I}_{\text{dir}}$ are the direction vectors for the text and image modalities, respectively. 

\noindent \textbf{Masked directional Loss.} As shown in~\cite{he2022masked}, randomly masking patches can speed up the reconstruction process and serve as a self-supervised training paradigm for image reconstruction tasks. In \net, the mask can work efficiently to enforce the convergence regarding style loss. Furthermore, we consider that the style in the images is locally consistent and independent from the contents. Hence, we mark partial contents to enforce the style similarity close to the complete image results. Therefore, we extend the directional loss with a masking step to incorporate robustness against partial visibility of style features. Given the content image $\textbf{X}$ and style text $t$, the stylized image $\textbf{Y}$ is generated and subsequently masked randomly by 50\% using $16 \times 16$ patches to obtain the masked image $Z$. The features of $Z$, $F_{I}(Z)$, and the content features $F_I(\textbf{X})$ are then computed to calculate the masked directional loss. The proposed masked directional loss $\mathcal{L}_{\text{md}}$ is defined as:
\begin{equation}
    \mathcal{L}_{\text{md}} = \frac{\|F_{I}(Z) - F_I(\textbf{X})\|}{\|F_{T}(t) - F_{T}(t_{\text{src}})\|}
    \label{eq:mask}
\end{equation}
\noindent Equation~\ref{eq:mask} leverages the reconstruction challenge as a self-supervised signal, enhancing the model's ability to maintain style fidelity even when significant portions of the image are masked, drawing insights from the effectiveness of high masking ratios in self-supervised learning settings.

\begin{figure}[h]
    \centering
    \includegraphics[width=\linewidth]{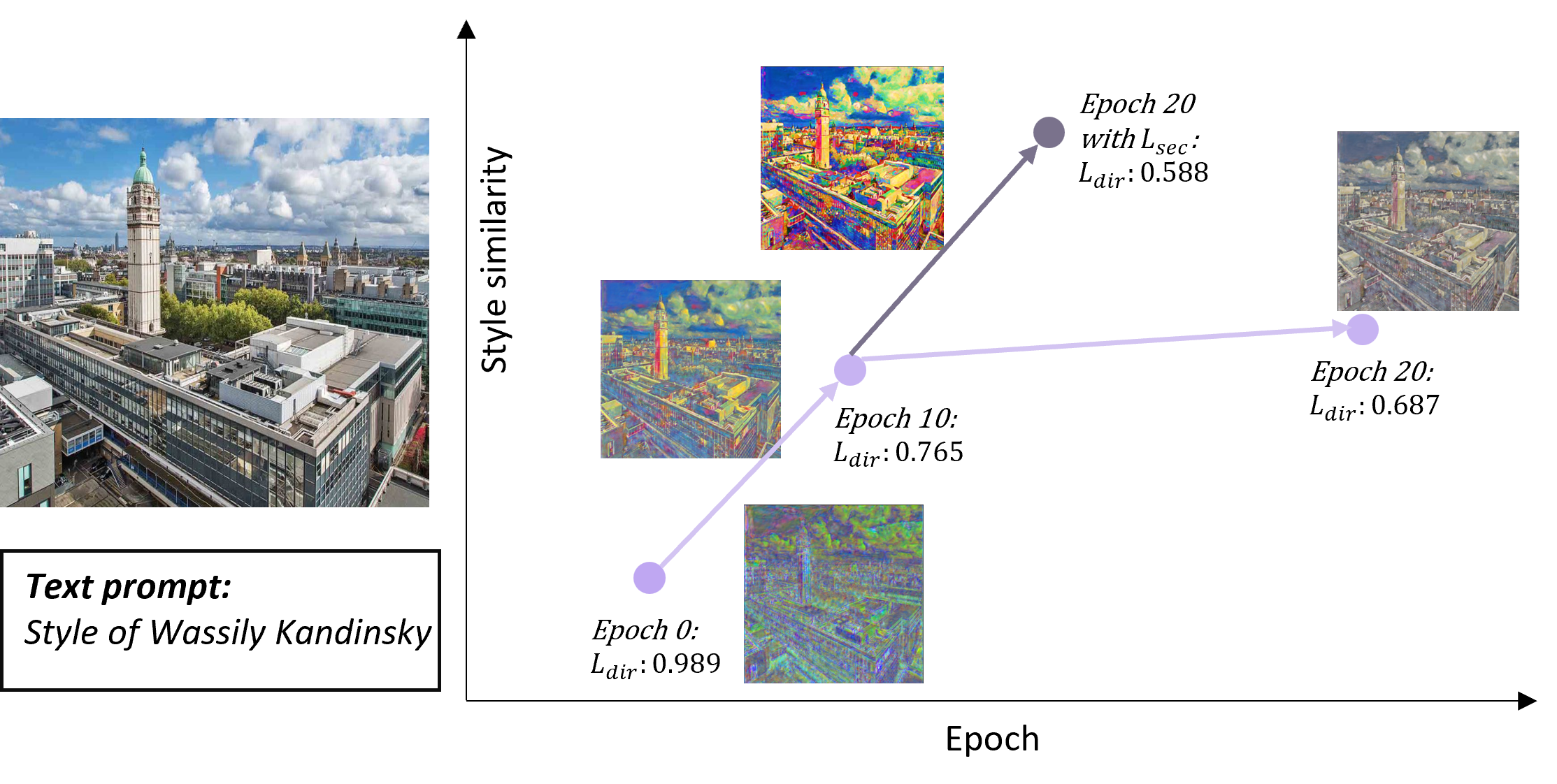}
    \caption{\textbf{Illustration of the proposed second-order directional loss.} It shows how $L_{so}$ allows for rapid adjustments in the direction of stylization. Notably, it facilitates refined stylistic shifts, ensuring a swift and coherent transition towards the desired visual style.}
    \vspace{10pt} 
    \label{fig:sec}
\end{figure}

\begin{figure*}[t]
    \centering
    \includegraphics[width=1\linewidth]{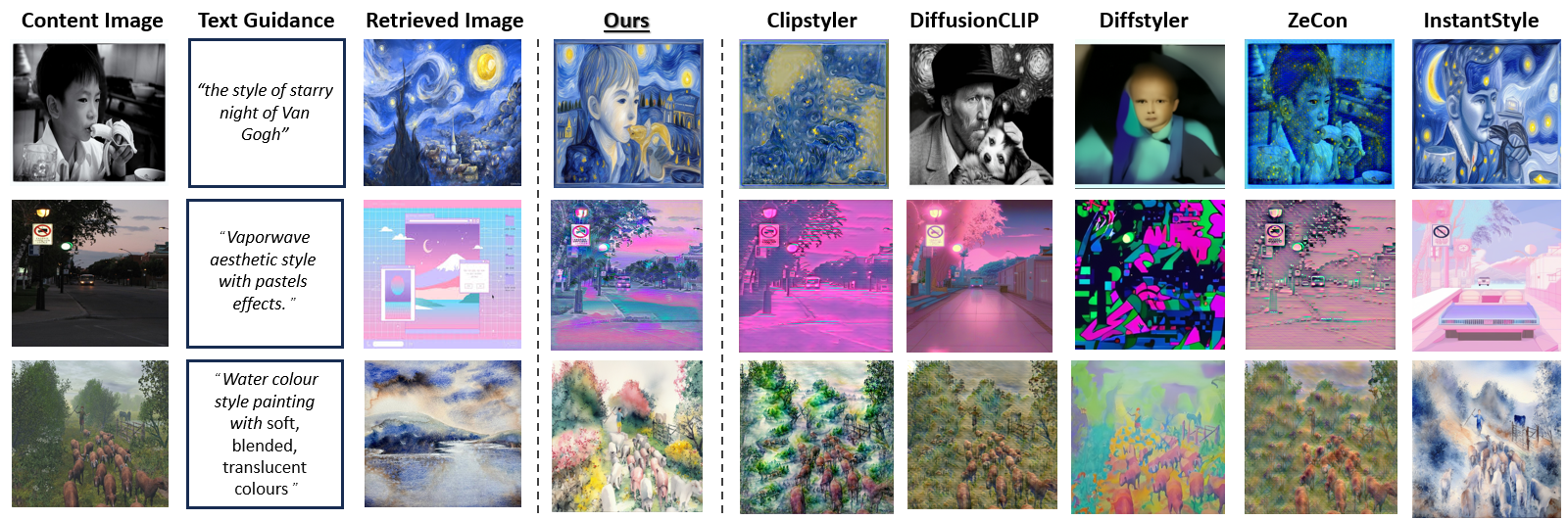}
    \caption{\textbf{Qualitive comparison with SOTA algorithms.} We show three cases of text-guided style transfer. For reference, we use the text to retrieve a reference image for style comparison and the input to InstantStyle.}
    \label{fig:sota_comp}
    \vspace{10pt}
\end{figure*}

\noindent \textbf{Second-order directional loss.} However, to speed up the convergence speed, the second-order loss $\mathcal{L}_{\text{so}}$ introduces a novel approach by considering the progression of the stylized image across successive epochs. As shown in Figure \ref{fig:sec}, the second-order directional loss ($L_{so}$) is designed to speed up the optimization of the alignment between the generated image and the textual style description. It achieves this by considering the changes in stylization direction across successive epochs, ensuring a coherent and swift transition towards the desired visual style. This is quantified by the squared norm of the difference between the image features at consecutive epochs, relative to the text direction vector, and is weighted by a dynamic term, ( $\alpha_{\text{shift}}$ ), to adjust the influence of the style shift on the loss function. Concretely, given a content image $\textbf{X}$, style text $t$, and stylized images $\textbf{Y}(i)$ and $\textbf{Y}(i+1)$ obtained at epochs $i$ and $i+1$, the second-order directional loss $\mathcal{L}_{\text{so}}$ is formulated as follows:

\begin{equation}
    \mathcal{L}_{\text{so}} = \left\|\frac{F_{I}(\textbf{Y}(i+1)) - F_{I}(\textbf{Y}(i))}{F_{T}(t) - F_{T}(t_{\text{src}})}\right\|^2 \cdot \alpha_{\text{shift}},
    \label{eq:so}
\end{equation}

\noindent which measures the change in stylization from epoch $i$ to $i+1$ relative to the change in text features. Specifically, $\alpha_{\text{shift}}$ is a dynamic weighting term that could be formulated as:

\begin{equation}
    \alpha_{\text{shift}} = \alpha \cdot \left(1 - e^{-\beta \|\mathbf{F}_{I}(\textbf{Y(i)}) - \mathbf{F}_{I}(\textbf{X})\|}\right)
    \label{eq:shift}
\end{equation}

\noindent where $\alpha$ and $\beta$ are hyperparameters that control the sensitivity of the loss to the magnitude of the style shift. Concretely, $\alpha$ serves as an upper limit or a scaling factor for the weight. It can be used to adjust the maximum influence the style shift weight can have on the second-order loss. $\beta$ is a scaling hyperparameter that controls the sensitivity of the weight to changes in the style feature difference. A larger $\beta$ would make the weight more responsive to smaller differences in style features. The exponential term ensures that the weight increases as the difference between the target and source style features grows, but it is bound to prevent the loss from escalating excessively. When the style feature difference is large (meaning the current style is far from the target), this term is close to 0, making $\alpha_{shift}=\alpha$, which in turn makes the weight larger, allowing the second-order loss to have a smaller impact on the training update, and vice versa. Furthermore, to enhance the training stability and ensure fast updates, a threshold is applied to the direction loss $\mathcal{L}_{\text{dir}}$. The second-order loss $\mathcal{L}_{\text{so}}$ is employed only if $\mathcal{L}_{\text{dir}}$ falls below a predefined threshold $\theta$, allowing for the refinement of style transfer in later stages of training. Including the aforementioned masked directional loss, the overall proposed text-driven style loss is defined as follows.

\begin{equation}
\mathcal{L}_{\text{style}} = \mathcal{L}_{\text{dir}} + \mathcal{L}_{\text{md}} + \mathbb{I}(\mathcal{L}_{\text{dir}} < \theta) \cdot \mathcal{L}_{\text{so}}
\label{eq:total}
\end{equation}

\noindent where $\mathbb{I}(\cdot)$ is the indicator function, which is 1 if the condition is true, and 0 otherwise.

\noindent \textbf{Content losses.} The VGG content loss \cite{sanet} ($L_{\text{vgg}}$) is used to supervise content similarity. It is defined as the sum of squared errors between the feature maps of the content and the stylized image, extracted from various layers of the VGG network. Meanwhile, we also use the LPIPS~\cite{LPIPS} loss $L_{\text{lpips}}$ to align the stylized results more closely with human perceptual judgments. 

\noindent \textbf{Total loss.} We incorporate the content loss and the style losses as $L_{total} = \alpha L_{\text{style}} + \beta L_{\text{lpips}} + \gamma L_{\text{vgg}}$, where $\alpha, \beta, \gamma$ are the weights that balance the contribution of each loss component to the total loss. This formulation ensures that the stylization aligns with the textual description while preserving the integrity of the content image.

\section{Experiments}
\label{sec:experiments}

\noindent $\bullet$ \textbf{Datasets.} To evaluate   our \net framework, we utilized two distinct datasets: COCO\cite{COCO} and WikiArt\cite{WikiArt}. It is important to note that these datasets were not employed during the training phase of our model. Instead, they are used in the testing and inference stages, providing a diverse range of images and artistic styles to assess the effectiveness of our text-driven image style transfer approach.  We use the COCO dataset as content images while applying various textual styles to evaluate the model's performance in capturing and rendering complex artistic styles.

\noindent $\bullet$ \textbf{Parameter setting.} For our experiments, we adapted the pretrained VAE from Stable Diffusion~\cite{ldm}, opting to keep the encoder fixed while only training the decoder. To ensure stable training, we introduced a two-stage learning rate strategy for the content loss weight. Initially, we set the weight at 9000, which is then reduced to 150 after 5 epochs. This approach was taken to prevent early overfitting to content features and to gradually refine the content alignment as the training progressed.  Additionally, our training procedure included a masking component, where we randomly masked 50\% of the patches. With these settings, we trained our models using the Adam optimizer with a starting learning rate of $5\times10^{-4}$, halving it at epoch 10, and terminating the training at 20 epochs. 

\noindent $\bullet$ \textbf{Metrics and evaluation.} To assessour textimage model's effectiveness in content consistency and style alignment, we employed three metrics: \textbf{1. CLIP score}: This is quantified by computing the cosine similarity between the features extracted by the CLIP (SigLIP) model from the text and the image. A \textbf{higher} CLIP similarity score indicates better style alignment. \textbf{2. SSIM}: It is used to measure the similarity between the content of the original image and the stylized image. A \textbf{higher} SSIM index signifies a better preservation of content. \textbf{3. VGG Loss}: It captures the content differences between content and stylized images by examining the feature responses at various layers of the VGG network. A \textbf{lower} VGG content loss implies better performance.

\subsection{Comparison with the state of the art}
\label{sec:sota}

\begin{table}[h]
\centering
\resizebox{\linewidth}{!}{
\begin{tabular}{l|c|c|c|c}
\hline
                     & CLIP score $\uparrow$ & SSIM $\uparrow$ & VGG loss $\downarrow$ & Aesthetic score $\downarrow$ \\ \hline
Clipstyler\cite{clipstyler}           & 0.180 & 0.988 & 2.106 & 5.379 \\
DiffusionCLIP~\cite{diffusionclip}        & 0.327 & 0.979 & 1.272 & 5.742 \\
Diffstyler~\cite{diffstyler}           & 0.351 & 0.982 & 1.182 & 4.660 \\
TxST~\cite{txst}                 & 0.409 & 0.990 & 0.975 & 4.552 \\
ZeCon~\cite{zecon}                & 0.417 & 0.971 & 1.218 & 5.327 \\
Our final model      & \textbf{0.492} & \textbf{0.996} & \textbf{0.547} & \textbf{4.172} \\ \hline
\end{tabular}
}
\caption{\textbf{Comparison with Sate-of-the-art methods.} \net is consistently better than other text-guided style transfer models regarding the CLIP score, SSIM, VGG loss and Aesthetic score.}
\label{tab:sota}
\vspace{10pt} 
\end{table}

To show the superiority of our proposed \net, we compare it with several state-of-the-art approaches, including DiffusionCLIP~\cite{diffusionclip}, Diffstyler~\cite{diffstyler}, TxST~\cite{txst} and ZeCon~\cite{zecon}. We first quantitively compare the performance of these models, the experiment is done on 10 random content images and 10 text prompts of various descriptions. Among them, TxST claims to be real-time stylization without online optimization, and ZeCon claims to be a faster version of Clipstyler~\cite{clipstyler}. The CLIP score of \net is 0.492, the highest in the group, indicating proficiency in generating semantically coherent images in response to textual descriptions. \net also shows higher values on SSIM and VGG content loss. These metrics underscore the model's ability to retain the structural integrity of the content images. Lastly, measured by the Aesthetic differences metric~\cite{aesthetic}, shows the lowest score among the compared models, suggesting that \net is capable of producing images that are aesthetically appealing to humans. Overall, our proposed \net effectively produces images that are visually aligned with text prompts, structurally similar to reference images and superior in aesthetic quality. A qualitative comparison is also conducted based on three cases as shown in Figure \ref{fig:sota_comp}. The retrieved image is derived based on the prompt text, which is used as a reference for the style transfer performance and the input to InstantStyle \cite{wang2024instantstyle}. Apparently, \net can achieve consistently better results in all cases regarding content preservation and style transfer ability. Concretely, \net outperforms Clipstyler, DiffusionCLIP, Diffstyler and ZeCon regarding both style transfer and content preservation capabilities. Compared with InstantStyle (\cite{wang2024instantstyle}), \net shows competitive style transfer ability without taking in a reference style image, while the content preservation ability is better than InstantStyle.

\begin{table}[]
\resizebox{\columnwidth}{!}{%
\begin{tabular}{lccc}
\hline
\multicolumn{1}{c}{Model} & \multicolumn{1}{l}{Training time}                      & \multicolumn{1}{l}{Inference Time}                   & \multicolumn{1}{l}{Model Param.} \\ \hline
Clipstyler\cite{clipstyler}                  & 403 sec & 4 sec & 28M                      \\
DiffusionCLIP \cite{diffusionclip}             &  293 sec & 56 sec                                               & 214 M                      \\
Diffstyler \cite{diffstyler}              & -                                                      & 42 sec                                               & 207M                      \\
TxST \cite{txst}                      & -                                                      & 27 sec                                               & 51M                       \\
ZeCon  \cite{zecon}                   & -                                                      & 36 sec                                               & 67M                       \\
Ours           & \textbf{23 sec}                                        & \textbf{2 sec}                                       & \textbf{17M}              \\ \hline
\end{tabular}%
}
\caption{\textbf{Computational complexity comparison of different models.} We record the training time, inference time, and the number of model parameters.}
\label{tab:speed_style}
\vspace{10pt} 
\end{table}

In Table~\ref{tab:speed_style}, we show the computation complexity among different approaches. We record both computation time and model parameters to demonstrate the efficiency of our proposed \net. Note that Diffstyler, TxST and Zecon require long offline training hours to achieve real-time stylization. The approximate training time could be hundreds of hours. We can see that ours achieves $10\sim 20\times$ speedup compared to Clipstyler and DiffusionCLIP on training time. Due to the proposed fast Mamba based style fusion module, \net also achieves a faster inference process, approximately 2$\times$ faster than Clipstyler.

\begin{figure*}[t]
    \centering
    \includegraphics[width=1\linewidth]{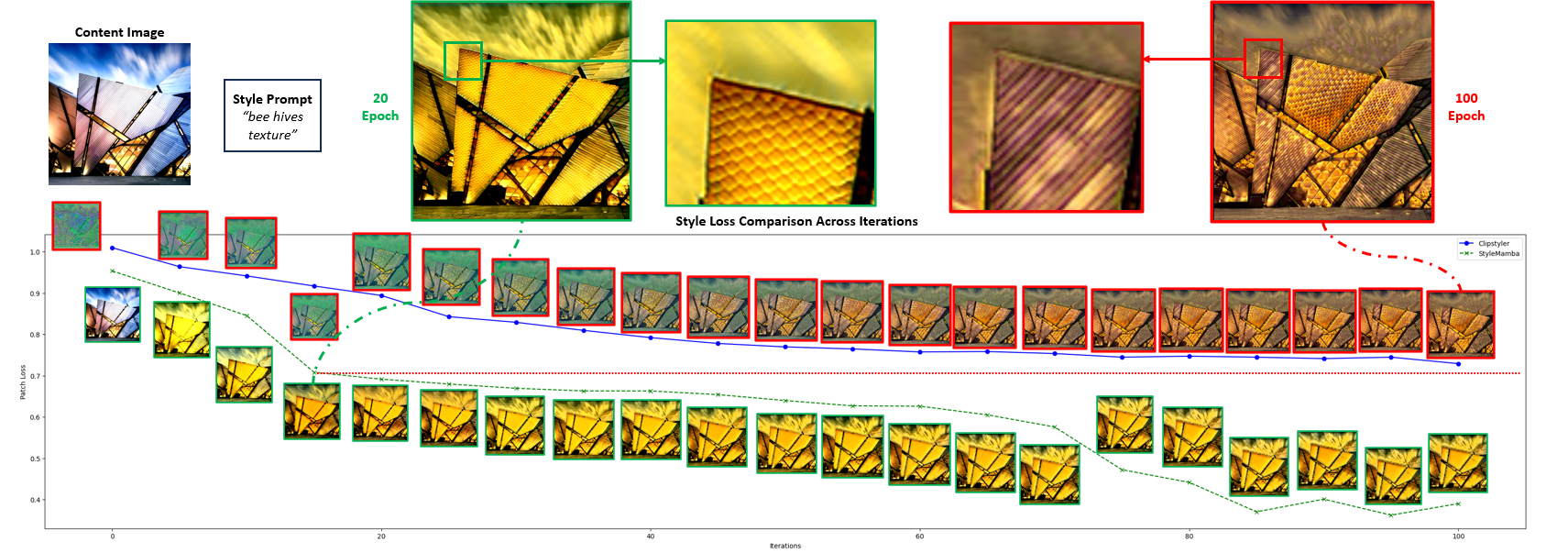}
    \caption{\textbf{Speed comparison between \net and Clipstyler.} The graph depicts the style loss and corresponding intermediate results over time, illustrating the convergence efficiency of each method. The result of \net at epoch 15 shows a more detailed stylization result than Clipstyler's 100 epoch result.}
    \label{fig:speed}
    \vspace{10pt} 
\end{figure*}

\subsection{Ablation study}
\label{sec:ablation}
\noindent \textbf{Style Fusion module.} To test the effectiveness of the style fusion module,  we compare two settings: ours with the Mamba module, and the cross-attention module used in \cite{txst}. Table \ref{tab:attention} shows the experiment results. The combination of SigLIP and Mamba achieves the best results regarding the CLIP similarity and the VGG content loss. We can also observe that using our proposed style fusion module achieves better stylization and  runs faster than the cross-attention module, approximately $3\times$ speedup. This is due to the efficiency of Mamba's sequential process over the cross-attention's global correlation. It can also be observed from the model complexity in Table~\ref{tab:speed_style}.

\begin{table}[]
\centering
\resizebox{\linewidth}{!}{
\begin{tabular}{l|ccccc}
\hline
Modules                & CLIP score $\uparrow$ & SSIM $\uparrow$ & Running time$\downarrow$ \\ \hline
Clipstyler                    & 0.180           & 0.988           & 4 sec                         \\
CLIP+cross-attention                & 0.387           & 0.979           &  7 sec                         \\
SigLIP+cross-attention              & 0.391           & 0.987           &  7 sec                         \\
CLIP+Mamba  & 0.422         & 0.990           &  2 sec                         \\
SigLIP+Mamba                  & \textbf{0.452}  & \textbf{0.991}  &  \textbf{2 sec}                \\ \hline
\end{tabular}
}
\caption{\textbf{Ablation study on Style fusion module.} We compare different Style Fusion modules used in the \net. The combination of SigLIP and Mamba achieves the best results regarding style transfer accuracy and speed.}
\vspace{10pt}
\label{tab:attention}
\end{table}

\noindent \textbf{Text-to-image models.} In Section~\ref{sec:approach}, we propose to replace CLIP~\cite{clip} with SigLIP~\cite{siglip} because SigLIP generally performs better in text-to-image alignment. Hence, it has the potential for a better understanding of style representation in the images. To demonstrate its efficiency, we randomly select 10 content images and 10 text prompts to test different models during the training and testing phases on the performance of text-image similarity, content preservation, and style transfer in the generated images. The average results are displayed in Table \ref{tab:clip}.   

\begin{table}[]
\centering
\resizebox{\linewidth}{!}{
\begin{tabular}{l|ccc}
\hline
Training \& evaluation                & CLIP score $\uparrow$ & SSIM $\uparrow$ & VGG loss $\downarrow$ \\ \hline
CLIP \& CLIP                  & 0.276           & 0.989           & 3.920                         \\
CLIP \& SigLIP                & 0.066           & 0.989           & 3.920                         \\
SigLIP \& CLIP                & \textbf{0.325}  & \textbf{0.990}  & \textbf{3.899}                \\
SigLIP \& SigLIP              & 0.127  & \textbf{0.990}  & \textbf{3.899}                \\ \hline
\end{tabular}
}
\caption{\textbf{Ablation study on text-to-image models.} To measure the alignment between texts and images, we use CLIP and SigLIP for model training and evaluation. The CLIP score is calculated by using corresponding evaluation models.}
\label{tab:clip}
\vspace{10pt}
\end{table}

In Table~\ref{tab:clip}, we show the training and evaluation combinations of using different text-to-image models. It shows that the experiments with the SigLIP model during evaluation (rows 3 and 4) outperform the corresponding experiments using the CLIP model (rows 1 and 2). Specifically, the CLIP similarity score increases from 0.276 to 0.325 and from 0.066 to 0.127, when comparing the CLIP testing model against the SigLIP testing model. This suggests that the SigLIP model is more effective at capturing and translating stylistic elements from text to image. Moreover, comparing rows 1 and 3, we can see that using SigLIP to train the model can significantly improve the CLIP score, which indicates that SigLIP is better at understanding the style representation and aligning the stylized image closer to the style prompts. This is also indicated by comparing rows 2 and 4.

\noindent \textbf{Content Encoder $F_e$.} Furthermore, we explored the synergistic effects of combining different text-to-image models with VGG and stable diffusion autoencoders (SD VAE)~\cite{ldm}. As shown in Table \ref{tab:style}, the results indicated that SD VAE outperforms VGG AE in conjunction with the same CLIP model, yielding a higher CLIP similarity score and lower VGG content loss. Specifically, we can see improvements with SD VAE (row 2 > row 1, and row 4 > row 3), suggesting that SD VAE is more effective in capturing and preserving style and contents. Further, the SigLIP model consistently delivered higher CLIP scores than CLIP models (1\&2, 3\&4), affirming the superiority of SigLIP in image-text alignment.

\begin{table}[]
\centering
\resizebox{\linewidth}{!}{
\begin{tabular}{l|ccc}
\hline
Modules                & CLIP score $\uparrow$ & SSIM $\uparrow$ & VGG loss $\downarrow$ \\ \hline
SigLIP+VGG AE                 & 0.322           & 0.990           & 3.874                         \\
SigLIP+SD VAE               & \textbf{0.349}  & \textbf{0.985}  & \textbf{2.759}                \\
CLIP+VGG AE                   & 0.126           & 0.789           & 3.852                         \\
CLIP+SD VAE                 & 0.140           & 0.785           & 2.937                         \\ \hline
\end{tabular}
}
\caption{\textbf{Ablation study on Encoder model.} To better preserve the contents, we test using different pretrained autoencoder (VGG~\cite{VGG} and SD VAE~\cite{ldm}) for style transfer.}
\label{tab:style}
\vspace{10pt} 
\end{table}

\noindent \textbf{Loss functions $\mathcal{L}_{\text{style}}$.} As shown in Table \ref{tab:loss}, to assess the impact of various losses on the stylization process, we conducted a series of experiments by systematically integrating different loss functions and evaluating their influence on stylization. For comparison, the baseline model only utilizes the global directional loss ($\mathcal{L}_{\text{dir}}$) and VGG content loss ($\mathcal{L}_{\text{vgg}}$) in row 1. Subsequently, in row 2, we enhanced the loss function framework of Clipstyler by incorporating the proposed masked directional loss ($\mathcal{L}_{\text{md}}$), aimed at refining the model's ability to capture directional attributes within the content image. This addition sought to bolster the CLIP score, a metric indicative of style-content alignment. Further augmentation, in row 3, involved the amalgamation of LPIPS ($\mathcal{L}_{\text{lpips}}$) with the masked directional loss($\mathcal{L}_{\text{md}}$), facilitating simultaneous improvements in both content preservation, as reflected by the content loss metric, and style alignment, as indicated by the improved CLIP score. Finally, in row 4, we integrated a second-order loss $\mathcal{L}_{\text{so}}$, implemented at an interval of every 5 epochs. It is pertinent to note that narrower intervals were avoided to prevent potential instability in the convergence pattern. The experimental outcomes, as presented in Table~\ref{tab:loss}, were generated under a consistent training duration of only 50 epochs across all models. 

Comparing rows 1 and 2, we can see that masked directional loss markedly enhances the style similarity. Rows 2 and 3 show that adding LPIPS loss can significantly improve content preservation, with approximately 2.2 reductions in VGG loss. The last row is the complete losses used in \net, we can see that adding second-order directional loss can further improve the style similarity without compromising the content information. 

\begin{table}[]
\centering
\resizebox{\linewidth}{!}{
\begin{tabular}{l|c|c|c}
\hline
 Loss terms   & CLIP score $\uparrow$ & SSIM $\uparrow$ & VGG loss $\downarrow$ \\ \hline
$\mathcal{L}_{\text{dir}}+\mathcal{L}_{\text{vgg}}$ (baseline)                       & 0.329           & 0.986           & 0.624                         \\
baseline+$\mathcal{L}_{\text{md}}$                        & 0.422           & 0.889           & 2.779                         \\
baseline+$\mathcal{L}_{\text{md}}+\mathcal{L}_{\text{lpips}}$                & 0.428           & 0.985           & 0.593                         \\
baseline+$\mathcal{L}_{\text{md}}+\mathcal{L}_{\text{lpips}}+\mathcal{L}_{\text{so}}$ & \textbf{0.475}  & \textbf{0.986}  & \textbf{0.572}                \\ \hline
\end{tabular}
}
\caption{\textbf{Ablation study on loss functions.} To compare different loss terms, we gradually add more loss terms to evaluate the performance. The final model combining $\mathcal{L}_{\text{md}}$, $\mathcal{L}_{\text{lpips}}$ and $\mathcal{L}_{\text{so}}$ achieves the best results in terms of the style similarity and content similarity.}
\label{tab:loss}
\vspace{10pt} 
\end{table}

To assess the training efficiency of the proposed \net compared to Clipstyler, we plot the progression of style loss over 100 epochs (Figure \ref{fig:speed}). The red line in the plot indicates that \net begins to exhibit signs of convergence by epoch 15 while Clipstyler achieves similar style loss at epoch 100. Approximately, we achieve 7$\times$ speedup. Qualitatively, we show the enlarged patches generated by the \net and Clipstyler, and we can see that ours can better transfer the ``\textit{bee hives}'' patterns to the content image. 

\begin{figure}[h]
    \centering
    \includegraphics[width=\linewidth]{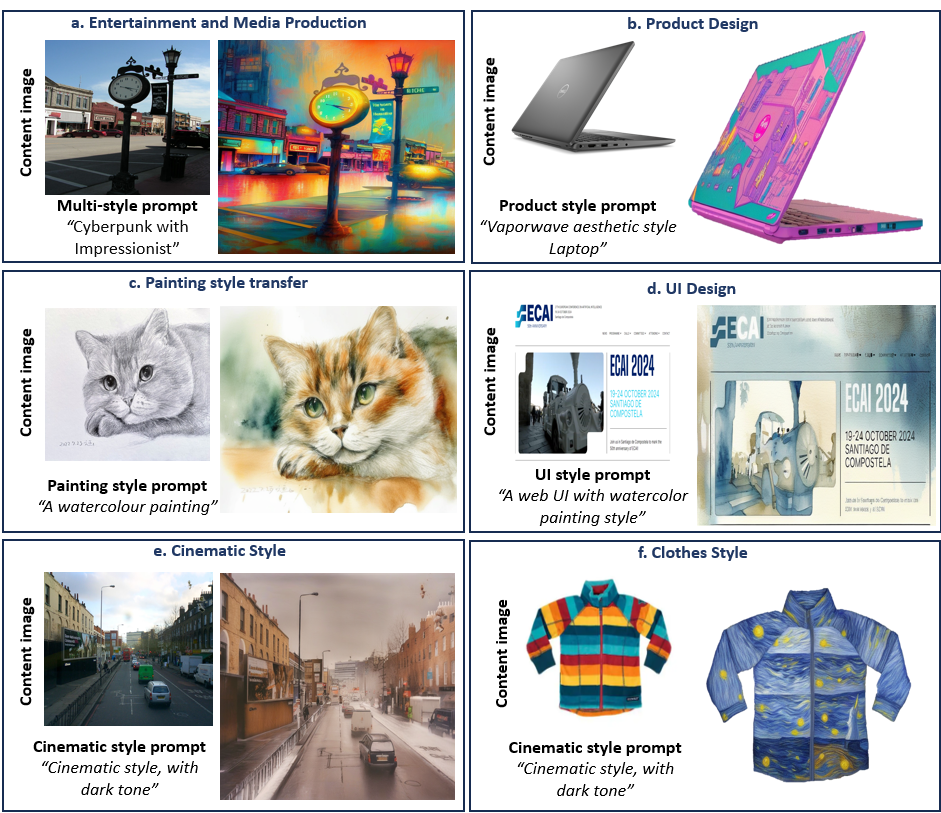}
    \caption{\textbf{\net in the wild}. Our better balance between content preservation and style transfer ability can enable many applications with high content preservation requirements, while with good style transfer ability.}
    \label{fig:wild}
    \vspace{10pt} 
\end{figure}

\subsection{Style transfer in the wild}
\label{sec:wild}


To test the generalization of our proposed \net, we show six real applications in Figure \ref{fig:wild}, which showcases the versatile capabilities of the proposed network across a range of creative domains, demonstrating its robust generalization ability. Figure~\ref{fig:wild} a shows the multiple style transfer. Given the two text prompts, ``\textit{cyberpunk}'' and ``\textit{impressionism}'', we see a coherent stylized result that can reflect both style features. Figure~\ref{fig:wild} b highlights the network’s application in product design. Knowing the mask of the contents, we can customize the appearance of the desktop by providing any text prompts. Figure~\ref{fig:wild} c shows that the proposed \net can understand the contents from sketches and generate desirable colors and textures to complete the painting, indicating its ability for painting assistance. This ability could speed up the process of art creation, providing artists with a powerful tool to visualize and experiment with different color palettes and textural effects. The network's prowess extends to user interface (UI) design, where in Figure~\ref{fig:wild} d, it reimagines the main page of ECAI 2024 with a watercolor painting style. This adaptation is not just cosmetic but conceptual, showcasing how a digital platform can embody the aesthetics of traditional art, potentially making digital experiences more organic and visually engaging. Then the cinematic style transformation in Figure~\ref{fig:wild} e highlights \net's capacity to alter not just the mood but the narrative tone of a scene. It adeptly applies a "dark tone" to a typical urban setting, demonstrating the potential to guide the viewer's emotional response and enrich tvisual media storytelling. . Lastly, Figure~\ref{fig:wild} f explores the frontier of fashion design. Here, \net demonstrates a keen eye for the transposition of iconic art onto wearable designs. The network takes the quintessential patterns of a painting and coherently wraps them around a three-dimensional garment, bridging the gap between classical art and modern apparel. This instance exemplifies how \net could be a transformative tool for fashion designers, enabling them to create innovative prints and textiles inspired by a vast array of visual art forms. 

These six applications not only affirm the adeptness of \net in understanding and executing complex visual prompts and illuminate its potential as a catalyst in creative industries, from enriching visual storytelling to revolutionizing product and fashion design.

\subsection{Limitations and discussion} 
\begin{figure}[h]
    \centering
    \includegraphics[width=1\linewidth]{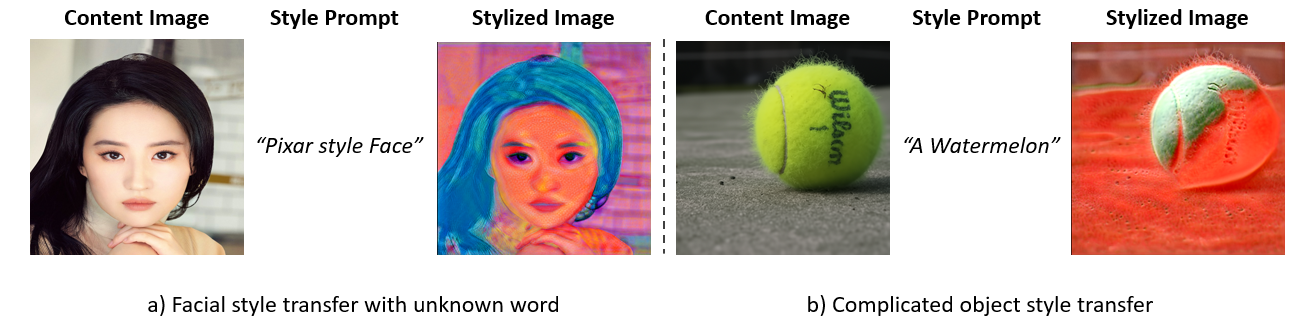}
    \caption{\textbf{Failure examples.} \net has some limitations on face images and object manipulation.}
    \label{fig:fail}
    \vspace{10pt} 
\end{figure}

As shown in Figure~\ref{fig:fail}, \net has limitations on understanding content-guided or less commonly used texts. For example, it can not understand ``\textit{Pixar styles}'' for face editing (Figure~\ref{fig:fail} a) or ``\textit{watermelon}'' (Figure~\ref{fig:fail} b) for shape transformation. Moreover, a controled style transfer can not be done with \net, while the CLIP doesn't have the segmentation capcity.

These limitations highlight areas for future research and development, such as improving the model's ability to handle diverse facial features and expanding its understanding of novel and abstract concepts for more accurate and varied style transfers.

\section{Conclusion}
\label{sec:conclusion}

This paper presented \net, an innovative framework for text-driven image style transfer that utilizes a conditional State Space Model integrated within an AutoEncoder architecture. Our framework demonstrates a significant advancement in the field by reducing the number of training iterations and also the inference time and training time required for each epoch, as well as enhancing the efficiency of the stylization process without sacrificing the quality of the stylized images. The results from our extensive experiments illustrate that \net outperforms existing baselines in terms of speed, stylization accuracy, and content preservation. The introduction of novel loss functions, masked and second-order directional losses, have been particularly effective in achieving high-quality stylization that aligns closely with textual descriptions. This capability allows for greater flexibility and creative expression in the applications of style transfer, making \net a valuable tool for both artistic endeavors and practical applications like UI design, clothes design, etc. In future work, we aim to explore the integration of more diverse linguistic inputs and expand the model's ability to handle a broader range of visual styles in a finer, controlled way. 






\bibliography{mybibfile}

\end{document}